\documentclass{article}
\usepackage[
    sglblindworkshop,
    final
]{neurips_2025}
\workshoptitle{AI for Music}

\usepackage{amsmath}
\usepackage{graphicx}
\usepackage[utf8]{inputenc} 
\usepackage[T1]{fontenc}    
\usepackage{hyperref}       
\usepackage{url}            
\usepackage{booktabs}       
\usepackage{amsfonts}       
\usepackage{nicefrac}       
\usepackage{microtype}      
\usepackage{xcolor}         
\usepackage{caption}
\usepackage{subcaption}
\title{Slimmable NAM: Neural Amp Models with adjustable runtime computational cost}

\author{%
  Steven Atkinson \\
  Atkinson Advanced Modeling, LLC\\
  \texttt{steve@atkinsonadvancedmodeling.com} \\
}

\begin{document}

\maketitle

\begin{abstract}
  This work demonstrates ``slimmable Neural Amp Models'', whose size and computational cost can be changed without additional training and with negligible computational overhead, enabling musicians to easily trade off between the accuracy and compute of the models they are using.
  The method's performance is quantified against commonly-used baselines, and a real-time demonstration of the model in an audio effect plug-in is developed.
\end{abstract}

\section{Introduction}
In recent years, Neural Amp Modeler (NAM)\footnote{\url{https://neuralampmodeler.com}} has seen widespread adoption for data-driven virtual analog modeling of musical equipment.
With this adoption, it has become the norm that musicians do not train the models they use.
Therefore, they are often unable to tailor the modeling process to suit the computational limitations of their use case.
For users who find that models are too CPU-intensive to be used in practical settings, model distillation \cite{polino2018model} is a common workaround.
However, without access to typical ML compute (e.g.\ a GPU), this is cumbersome and/or time-consuming, and it disrupts creative workflows.
By contrast, finite impulse responses (IRs)---a ubiquitous data-driven model for musicians---can be easily made more computationally-lightweight by truncation to an arbitrary length, allowing musicians to trade off accuracy for compute on the fly.
One might reasonably desire a similar option in the realm of neural modeling.

This work demonstrates introduces ``slimmable NAMs''---neural networks whose size (and, therefore, computational cost) can be reduced after training with negligible computational cost.
We focus on NAM's stacked WaveNet architecture due to its strong performance, though the method can be adapted to NAM's other open-source architectures in a similar way.
A slider control is introduced to the graphical user interface (GUI) of an audio effect plugin that controls the size of the model in real-time, allowing for its user to easily explore the trade-off between compute and accuracy.

\section{Method}
A WaveNet \cite{van2016wavenet} is a multi-layer convolutional neural network whose layers operate on a time series of $c$-dimensional vectors. 
To slim the network from a width of $c$ to $c'$, the weights $\mathbf W \in \mathbb R^{c \times c \times k}$ and biases $\boldsymbol b \in \mathbb R^c$ of each convolutional layer are truncated, yielding $\mathbf W' \in \mathbb R^{c' \times c' \times k}$ and $\boldsymbol b' \in \mathbb R^{c'}$, respectively. 
For the affine projections connecting the inputs of dimension $d_x$ and outputs of dimension $d_y$ to the network ($d_x=d_y=1$ for mono audio data), only the rows of the input projection and columns of the output projection are truncated so that the input and output dimensionality is preserved.

NAM trains models through supervised learning using a ``dry/wet'' pair of audio files.
During training, in each mini-batch, the network is slimmed to a randomly-chosen $1 \le c' \le c$, and the slimmed network's predictions are supervised against the targets.
The resulting neural network can be used with any number of channels when making predictions.
Open-source code is available for training and real-time prediction \cite{Atkinson_SlimmableNamTrain_2025,Atkinson_SlimmableNamDsp_2025}, and installers are available for a plugin that integrates the latter \cite{atkinson2025gateway}.

\section{Results}
\begin{figure}[htbp]
    \centering
    \begin{subfigure}[b]{0.5\textwidth}
        \centering
        \includegraphics[width=\textwidth]{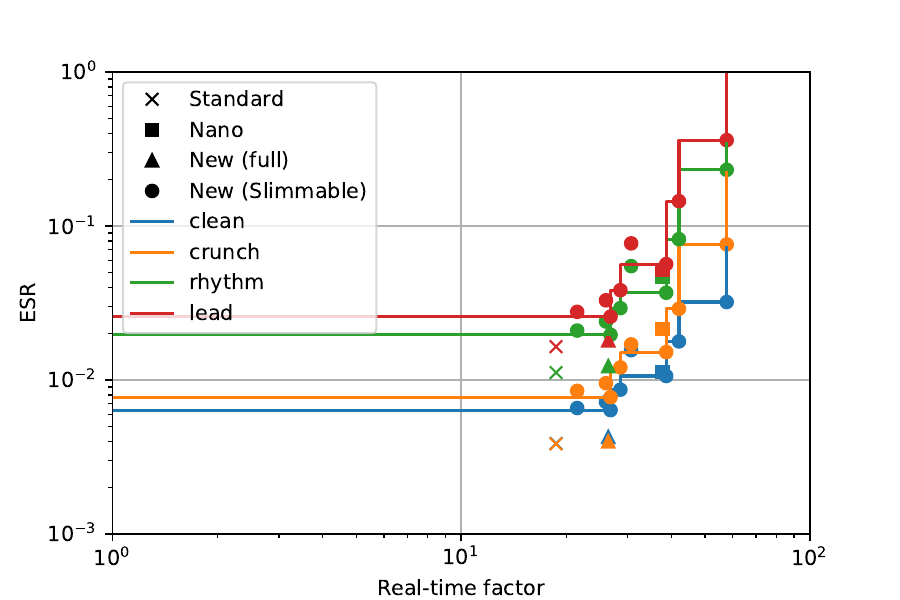}
        \caption{}
        \label{fig:metrics}
    \end{subfigure}
    \hfill
    \begin{subfigure}[b]{0.45\textwidth}
        \centering
        \includegraphics[width=\textwidth]{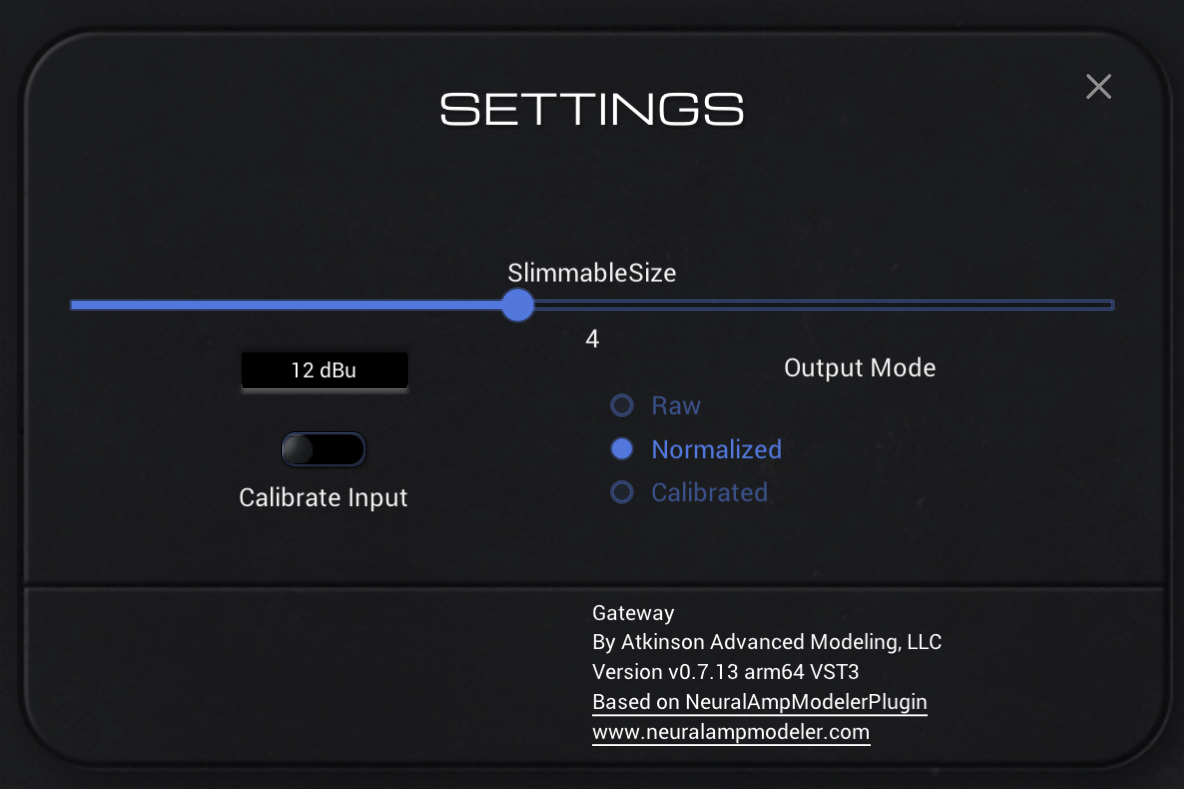}
        \caption{}
        \label{fig:plugin}
    \end{subfigure}
    
    \caption{
      \subref{fig:metrics}: Real-time factor (higher is better) and accuracy (error-signal ratio \cite{wright2020real}; lower is better) for different models of the four tones considered.
      The Pareto front is drawn for the slimmable NAM.
      ``New (full)'' refers to the a model using the slimmable architecture trained only for use at full size as normal.
      \subref{fig:plugin}: Plugin settings page, showing the ability to slim the loaded network on the fly.}
    \label{fig:main}
\end{figure}

Models were trained on recordings of guitar amplifiers spanning a range of tones: a ``clean'' tone recorded from a Fender Deluxe Reverb, a ``crunch'' tone recorded from a Morgan MVP23, and hi-gain ``rhythm'' and ``lead'' tones recorded from an Omega Ampworks Obsidian.
Figure \ref{fig:metrics} shows the trade-off between compute and accuracy for the slimmable NAM alongside other commonly-used models.
For audio examples as well as links to model files, see the video demonstration.\footnote{\url{https://youtu.be/93WAQsFu694}} 
Instead of using NAM's ``standard'' architecture, which\footnote{As of \url{https://github.com/sdatkinson/neural-amp-modeler/releases/tag/v0.12.0}} stacks two WaveNet models with different widths in series, this demonstration uses a new architecture with a single WaveNet module for simplicity. 
Fig.\ \ref{fig:plugin} shows the settings page of the plugin, adapted from NeuralAmpModelerPlugin\footnote{\url{https://github.com/sdatkinson/NeuralAmpModelerPlugin}}.
Moving the slider adjusts $c'$, allowing the user to audition models of different sizes in real time.

\section{Conclusions}
This work demonstrates Slimmable NAMs and demonstrates their performance and practical use as musical tools.

Decreasing the computational cost of a neural network without additional training is not new; examples include slimmable neural networks \cite{yu2019slimmable} which limit the width of the network as in our work; as well as dynamic gating and conditional computation \cite{ehteshami2020dynamic}; multi-exit and early-exit networks \cite{saeed2022binary}; and supernets \cite{cai2020once}.
In audio, Slimmable neural networks have already been applied in speech applications \cite{elminshawi2025dynamic}.
In virtual analog modeling, pruning has been explored for reducing the computational cost of neural models \cite{sudholt2022pruning}, though this requires additional training.
Slimmable models have value in NAM due to their ability to simultaneously remove the compute barrier to real-time prediction while simultaneously placing no additional restrictions or complications to the model-fitting process, making them an attractive solution for both model-makers and the musicians who play them. 
The slimmable WaveNet architecture introduced by this work is under investigation as the next-generation default model architecture for NAM, with the hope that it will increase the accessibility of the project to musicians.

\bibliographystyle{unsrt}
\bibliography{references}

\end{document}